\documentclass[a4paper]{article}
\usepackage[T1]{fontenc}
\usepackage{INTERSPEECH2021,float,subcaption}
\title{Insights on Neural Representations for End-to-End Speech Recognition}
\name{Anna Ollerenshaw$^1$, Md Asif Jalal$^1$, Thomas Hain$^1$}
\address{
  $^1$Speech and Hearing Research Group, University of Sheffield, UK}
\email{alollerenshaw1@sheffield.ac.uk}

\begin{document}

\maketitle
\begin{abstract}
  End-to-end automatic speech recognition (ASR) models aim to learn a generalised speech representation. However, there are limited tools available to understand the internal functions and the effect of hierarchical dependencies within the model architecture. It is crucial to understand the correlations between the layer-wise representations, to derive insights on the relationship between neural representations and performance. 
  Previous investigations of network similarities using correlation analysis techniques have not been explored for End-to-End ASR models. This paper analyses and explores the internal dynamics between layers during training with CNN, LSTM and Transformer based approaches using Canonical correlation analysis (CCA) and centered kernel alignment (CKA) for the experiments. It was found that neural representations within CNN layers exhibit hierarchical correlation dependencies as layer depth increases but this is mostly limited to cases where neural representation correlates more closely. This behaviour is not observed in LSTM architecture, however there is a bottom-up pattern observed across the training process, while Transformer encoder layers exhibit irregular coefficiency correlation as neural depth increases.
Altogether, these results provide new insights into the role that neural architectures have upon speech recognition performance. More specifically, these techniques can be used as indicators to build better performing speech recognition models.
\end{abstract}

\section{Introduction}

Traditionally, ASR frameworks have been developed using Hidden Markov Models (HMM) in combination with Gaussian Mixture Models (GMM) to identify and map acoustic features to phonemes. Recent work \cite{graves2013speech} has introduced deep neural networks to replace traditional approaches by factorising the system into specialised modules, such as acoustic and language models.
The End-to-End approaches for ASR attempt to simplify the pipeline and directly model the input features to characters or phonemes \cite{graves2014towards}. This approach allows the development of a complete ASR system without the requirement of expert domain knowledge, while attempting to globally optimise the training process.
As the development and integration of End-to-End approaches have become increasingly popular, many different architectures have been developed \cite{miao2015eesen} \cite{amodei2016deep}
\cite{bahdanau2016end} \cite{watanabe2018espnet}. Research from \cite{montufar2014number} has shown that neural layer depth can attribute to a richer neural representational capacity, but generalisation or memorisation behaviour of the models remains elusive \cite{arpit2017closer}. This hypothesis does not always translate to performance improvements in all cases \cite{basha2020impact} and has little exploration in the End-to-End ASR domain. End-to-End architectures have inherently complex internal dynamics and whether the model learns to generalise from the training process, is imperative to yield recognition performance improvements \cite{furui2009generalization} \cite{ogawa2015asr}. Furthermore, there is limited research with regard to the interactions between the training dynamics of End-to-End models and speech data and it is unclear how structural components or residual connections within the models contribute to more optimal representations. To explore this further, a window of observation into the neural representations of the network architectures would be required to provide information on the interaction between the training structures and the data. 

Current correlation analysis techniques from \cite{raghu2017svcca} and \cite{kornblith2019similarity} have been utilised to compare deep neural network representations. Comparing population representations has been explored in several methods, however this is a non-trivial task as it not clear which aspects of the representations the similarity index should attempt to focus on. Canonical Correlation Analysis (CCA) \cite{hardoon2004canonical} and Centered-Kernel Alignment (CKA) \cite{kornblith2019similarity} have been used as tools to compare network representations, as they enable the identification of shared structures across representations which are trivially dissimilar. The application of singular value decomposition (SVD) before CCA, referred to as SVCCA \cite{raghu2017svcca} has been used to compare representations across networks and it was found that network solutions for image classification, diverged predominantly in the intermediate neural layers. SVCCA has also been used as a tool to show the evolution of linguistic features as they were encoded in language models \cite{saphra2018understanding} and it has been observed, for an image classification task, that as neural layer depth increased, the network similarities decreased. Finally, the CKA approach demonstrated that task trained neural layers developed more similar representations than layers that were randomised.

However, these approaches have not yet been used to analyse neural representations of End-to-End architectures with speech data for speech recognition. In this work, a comparative study of neural representation analysis is provided with some of the most prevalent End-to-End ASR networks. The analysis focuses on understanding the similarity of the neural representations and answering questions such as: given the same input, how similar are the learned representations across training? Which model architectures have the highest impact upon similarity? How does neural similarity correspond to model performance?
Using correlation analysis techniques from \cite{kornblith2019similarity} and \cite{raghu2017svcca}, the key contributions are summarised as follows:
\begin{itemize}
    \item Development of a framework for correlation analysis across neural representations within state-of-the-art End-to-End architectures, Section \ref{sec:3.1}.
    \item A comparative analysis of similarity indexes upon an ASR task for different architectures, Sections \ref{sec:3.2} and \ref{sec:3.3}.
    \item Verification that internal representation analysis of End-to-End network structures can be used to magnify pathological components throughout model training, Section \ref{sec:3.3}.
    \item Discussion of the distinctions regarding the representations of End-to-End architectures and possible further work these observations can contribute to, Section \ref{sec:4}.
\end{itemize}

\section{Similarity Indexes for End-to-End ASR}\label{sec:2}

The End-to-End ASR task is to identify the acoustic input sequence $X = \{x_1,...,x_T\}$ of length $T$ as a label sequence $Y = \{y_1,...,y_N\}$ of length $N$ and directly map to the posterior distribution $p(Y|X)$. Due to the undefined separation of modules within End-to-End ASR networks, it is relatively unclear which, what and where the traditionally separate ASR system tasks are occurring, such as acoustic or language modelling. The internal parameter dependencies upon the structures of the model, and their effect upon the resulting performance, are ambiguous and inherently complex.

Using statistical correlation analysis methods, it is possible to relate two sets of observations within a network to find their correlation relationship. For the dataset $X = \{x_1,...,x_N\}$ and neuron $i$ in layer $l$, the activation output vector $z_i^l=(z_i^l(x_1),...,(z_i^l(x_N))$. By conducting correlation analysis techniques that are invariant to affine transforms, this enables comparisons between different neural networks and observations on the dynamic behaviour.

\subsection{SVCCA}\label{sec:2.1}

SVCCA \cite{raghu2017svcca} is used to find bases $w, s$ for two matrices such that, when the original matrices are projected onto these bases, their correlation is maximised:
\begin{equation}
    \frac{w^T\sum_{XY}s}{\sqrt{w^T \sum_{XX}w}\sqrt{s^T \sum_{YY}s}}
\end{equation}
where $\sum_{XX}, \sum_{XY}, \sum_{YY}$ are the covariance and cross-covariance. In the case of ASR neural networks, this is between the neural layers for $N$ data points where $l_1=\{z_1^{l1},...,z_{N1}^{l1}\}$ and $l_2=\{z_2^{l2},...,z_{N2}^{l2}\}$. The projected views of $l_1$ and $l_2$ are reduced to the top 99\% representative dimensions, using SVD, in an attempt to reduce potential noise in the representations, to form subspaces $l_1' \subset l_1, l_2' \subset l_2$. 
CCA \cite{thompson2005canonical} is then used to maximise the correlation of the projections of the linear transform of $l_1',l_2'$ by identifying vectors $w, s$ to maximise:
\begin{equation}
    \rho = \frac{\langle w^T l_1',s^Tl_2'\rangle}{||w^T l_1'||\; ||s^T l_2'||}
\end{equation}
The correlations of $\rho$ are higher when the representations have encoded more similar information.


\subsection{CKA}\label{sec:2.2}

CKA, first introduced in \cite{shawe2002kernel}, resembles CCA but is weighted by the eigenvalues of the corresponding eigenvectors. It is also similar in effect to SVCCA but incorporates the weighting symmetrically and doesn't require matrix decomposition. Instead of comparing multivariate features of the neural layers, the coefficiency between every pair of examples in each representation is measured, then the correlation computation is conducted. To measure the similarity index between the internal representations, the inner product is taken:

\begin{equation}\label{eq:crossprod}
    \langle vec(l_1l_1^T),vec(l_2l_2^T) \rangle = tr(l_1l_1^Tl_2l_2^T)=||l_2^Tl_1||^2_F
\end{equation}

where the elements of $l_1l_1^T$ and $l_2l_2^T$ are dot products between neural representations $z(x_1),...,z(x_T)$, using calculations from \cite{kornblith2019similarity}. 
The first half of Equation \ref{eq:crossprod} measures the similarity between examples while the second half has the same result by measuring between features by taking the sum of the squared dot products between every pair. For centered $l_1,l_2$:

\begin{equation}\label{eq:cov}
    \frac{1}{(n-1)^2}tr(l_1l_1^Tl_2l_2^T)=||cov(l_1^T,l_2^T)||_F^2
\end{equation}

The Hilbert-Schmidt Independence Criterion (HSIC) \cite{gretton2005measuring} generalises Equations \ref{eq:crossprod} and \ref{eq:cov} to the inner products from the kernel spaces, where the squared Frobenius norm of $\sum_{XY}$ becomes the squared Hilbert-Schmidt norm of the operator. This is equivalent to calculating a distance covariance. Where $K_{ij} = k(x_i,x_j)$ and $L_{ij}=l(y_i,y_j)$ where $k$ and $l$ are kernels, the estimator of HSIC is defined as:

\begin{equation}
    \mathit{HSIC}(K,L)=\frac{1}{(n-1)^2}tr(KHLH)
\end{equation}

where $H$ is the centering matrix $H_n=I_n-\frac{1}{n}11^T$. When $HSIC=0$, this suggests independence when $k$ and $l$ are universal kernels. However HSIC is not invariant to scaling until it has been normalised:

\begin{equation}
    \mathit{CKA}(K,L)=\frac{\mathit{HSIC}(K,L)}{\sqrt{\mathit{HSIC}(K,K)\mathit{HSIC}(L,L)}}
\end{equation}

These properties limit the use of CKA to be conducted across features rather than examples, for large models due to the size and the dramatically large computational costs that would be required.

\section{Experiments \& Results}\label{sec:3}


\subsection{Experimental Framework}{\label{sec:3.1}}

To analyse the internal representations of the models on an End-to-End ASR task, the experiments were done using the ESPRESSO framework \cite{wang2019espresso}. Each network was trained using the Switchboard dataset \cite{godfrey1992switchboard} with 300 hours of transcribed speech. This enabled the development of consistent state-of-the-art architectures and environment variables for End-to-End ASR. 


To investigate the time dependencies of the neural representations within End-to-End models, SVCCA and CKA, described in Section \ref{sec:2}, were applied to the activation outputs of each model layer across time. In order to conduct a comparable correlation analysis for each network and analysis method, several further steps were necessary: firstly, the network models were preserved at each epoch of the ASR task; and then they were fed to a separately developed pipeline for the extraction of the activation embeddings for each neuron. To ensure consistency, this was done by feeding in a controlled input of 100 speech frames to all architectures and extracting the activation output at each neuron, enabling the representation analysis methods to be conducted concurrently. Linear interpolation of the narrower layer to the same dimensionality as the wider layer was conducted due to the different spatial dimensions of the neural layers and thereby data-points, as both SVCCA and CKA methods require representation vectors to be the same dimensions. To compare the coefficiency correlation across the number of layers in the network, the spatial dimensions of the activation outputs were flattened into the number of data-points, in order to provide a spatial representation of each data-point.

\subsection{Encoder-Decoder Neural Representation Analysis}{\label{sec:3.2}}

The convolution network used in this work is a multi-layer stacked 2-dimensional convolution, with kernel size (3,3) on both the feature and the time axis from \cite{zhang2017very}. 
The final layer of the sequence-to-sequence model is then projected to an LSTM decoder from \cite{bahdanau2014neural}, and context at each time-step is generated with Bahdanau attention \cite{chorowski2015attention}. The encoder-decoder model function can be described by:
\begin{equation}
    \hat{y}^{dec}_{u} = LSTM(c_{u},y_{u-1},\hat{y}^{dec}_{u-1})
\end{equation}
where $\hat{y}$ is the hypothesised output by the model, $c_u$ is the context vector obtained by the encoder output, calculated by the attention mechanism.

\begin{figure}[t]
    \centering
    \includegraphics[width=\linewidth]{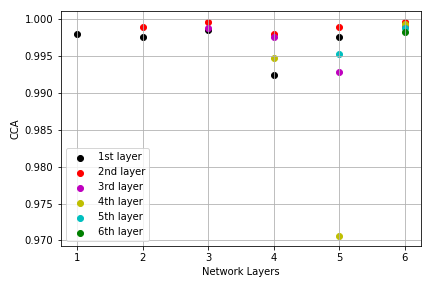}
    \caption{The final converged correlation coefficiency across all models with ascending CNN layers}
    \label{fig:all_cnn}
\end{figure}

This architecture allowed the observation of the layer-wise representation analysis methods across scaled convolutional layers within an ASR task. Comparison across layers allows the observation of the converged layer correlations, while comparison across epochs shows the hierarchical representations within the layers as the models train. Upon evaluation with the Hub5'00 set \cite{sundaram2000isip}, the word error rate performance of this architecture is displayed in Table \ref{tab:cnn width}. Increasing the neural depth improved accuracy slightly up to 3 layers but results varying the spatial dimensions of each layer showed little improvement. The performance was observed to be limited when varying the amount of neurons in each layer across the variable sized CNN architectures, with the best WER performance achieved with a 3 layer CNN. Upon comparing these coefficiency correlations across the CNN architectures with varied layers, shown in Figure \ref{fig:all_cnn}, the architectures with deeper spatial dimensions have more variation in the neural representations than the architectures that had the greater performance results, as suggested by results in \cite{kornblith2019similarity}, except for the 6 layer CNN model which had similar coefficiency but worse performance than the other models; this would require further investigation. 

\begin{table}[th]
  \caption{Variable sized CNN layers for End-to-End ASR evaluated on the Hub5'00 test set}
  \label{tab:cnn width}
  \centering
  \begin{tabular}{ l@{}c  c }
    \toprule
    \multicolumn{1}{c}{\textbf{CNN Architecture}} & 
                                         \multicolumn{1}{c}{\textbf{SWBD WER\%}} & \multicolumn{1}{c}{\textbf{Clhm WER\%}} \\
    \midrule
    6 layers & 11.4 & 22.4 \\
    5 layers & 10.7 & 21.3 \\
    4 layers & 10.9  & 21.2 \\
    3 layers & \textbf{10.5} & \textbf{20.8} \\
    2 layers & 10.6 & 20.9 \\
    1 layer & 11.6 & 22.5 \\
    \bottomrule
  \end{tabular}
\end{table}

\begin{figure}[t]
  \centering
  \includegraphics[width=\linewidth]{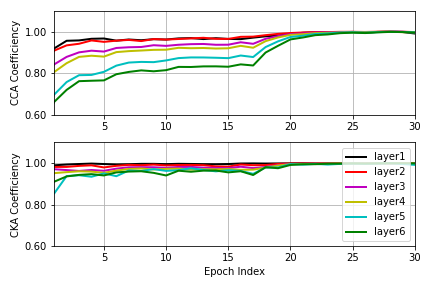}
  \caption{CNN neural representations evaluated with SVCCA (\textbf{top}) and CKA (\textbf{bottom}) through time as performance converges}
  \label{fig:CNN}
\end{figure}

As the number of the layers increased, the coefficiency of each layer approached 1, as shown in Figure \ref{fig:CNN}. Using SVCCA analysis, described in Section \ref{sec:2.1}, to correlate the activations across the epochs, it was observed that layers 1, 2 and 3 converge together at epoch 17, whereas deeper layers (layers closer to the output) converged slightly later but at the same point in training. 

Figure \ref{fig:CNN} also shows the CKA coefficiency, described in \ref{sec:2.2}, of the CNN architecture, where it can be generally observed that the SVCCA analysis is more sensitive to the initialisation parameters than CKA. With both strategies, a hierarchical correlation within the layers across training can be observed, although the CKA results suggest that there is some pathological behaviour present in deeper layers; for example the small spikes in coefficiency across layer 6. The CKA results potentially differ from the SVCCA results, due to the pruning of the SVD component of SVCCA while also assuming that all the coefficiency vectors are equally important to the representation of the ASR task.

\begin{figure}[t]
    \centering
    \includegraphics[width=\linewidth]{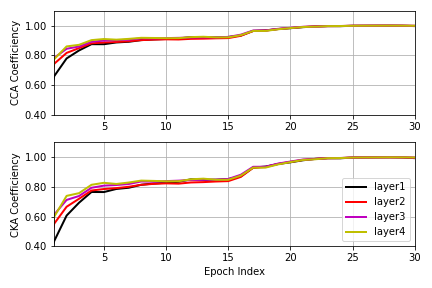}
    \caption{LSTM correlation coefficients of neural representations through time as performance converges  with SVCCA (\textbf{top}) and CKA (\textbf{bottom})}
    \label{fig:lstm}
\end{figure}

The LSTM neural representations, within the encoder-decoder framework, are displayed in Figure \ref{fig:lstm}. Comparing the SVCCA correlation results with the CKA results, it can be observed that the variance is slightly under-estimated by the SVCCA implementation, although both techniques display similar attributes. By comparing the internal representations with SVCCA and CKA, the behaviour of the internal neural dynamics of the architecture can be observed to be invariant to transformations, in a robust method. The coefficiency across epochs suggests that there is a bottom-up behaviour within the LSTM representations, with convergence occurring around epoch 22.

\subsection{Transformer Neural Representation Analysis}{\label{sec:3.3}}

The Transformer architecture from \cite{vaswani2017attention} was trained using the same dataset, with all 12 encoder blocks containing identical spatial widths. Due to the size of this architecture, the training was conducted across 90 epochs, to ensure model convergence. The Transformer model uses stacked self-attention and point-wise, fully connected layers for the encoder and decoder. Each block has a multi-head self-attention layer and feed forward layer. To analyse the representations of the Transformer encoder layers, the representations were unrolled across time steps. 

\begin{figure}[t]
    \centering
    \begin{subfigure}[t]{\linewidth}
        \includegraphics[width=\linewidth]{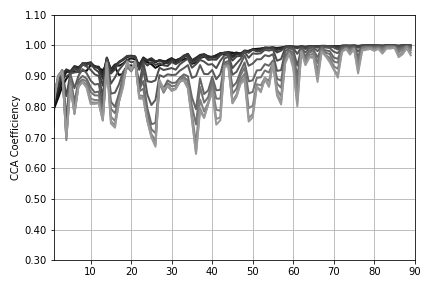}
    \end{subfigure}
    \begin{subfigure}[t]{\linewidth}
        \includegraphics[width=\linewidth]{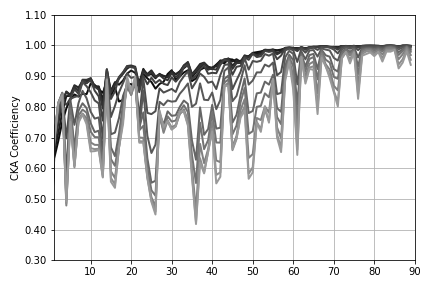}
    \end{subfigure}
    \caption{Transformer correlation coefficients through time as performance converges produced with SVCCA (\textbf{top}) and CKA (\textbf{bottom}), the darker colour gradients are higher layer representations, while the lighter the gradient the deeper the layer}
    \label{fig:transformer}
\end{figure}

The Transformer layer encoder output representations, shown in Figure \ref{fig:transformer}, emphasise the attending mechanism pathology present after the self-attention and linear operator. It can be observed that the higher layers of the Transformer encoder are less susceptible to the attention pathology than the deeper layers, which don't converge smoothly even after 80 epochs. There is a more noticeable distinction that can be ascertained from the CKA analysis, which retained more emphasised results, that there is similar overall hierarchical learning dynamics as observed in the CNN architecture.



\section{Discussion} \label{sec:4}

By using SVCCA as a method of analysing the internal representations for an End-to-End ASR framework, a window is observed on the dynamics of the training behaviour, although the top vectors pruned appear to under-represent the neural representations compared to using CKA. This is partly due to the assumption that all of the CCA vectors are equally important to the neural representation but also the SVD component of the SVCCA technique in Section \ref{sec:2.1} relies on the reflection of class information, which, for End-to-End speech recognition, is a potential limitation. By implementing the CKA analysis method in Section \ref{sec:2.2} it is possible to visualise the pathology of neural representations during training, particularly in the Transformer model, Figure \ref{fig:transformer}, which is indicative of the attention mechanism augmenting context information, particularly for wider layers. 


The techniques described in Section \ref{sec:2}, allow the observation of hierarchical behaviour of CNN and Transformer neural representations across training, Figures \ref{fig:CNN} and \ref{fig:transformer}, whilst also providing insight on the bottom-up invariant behaviour dynamics within the LSTM layers (without residual connections), shown in Figure \ref{fig:lstm}. The learning dependencies between layers across time exhibit similar learning dynamics as language models \cite{mittal2020learning}. The similarity indexes could also be used in future work to compare the correlation of the trained neural layers of various architectures across different speech datasets, such as noisy or augmented data, to observe how the neural layers respond dynamically during the training process. These experiments could then be directly correlated with the performance results.

Additionally, it has been noticed that scaling the depth of the convolutional layers had a limited effect upon network performance also in the case of End-to-End ASR, as shown in Table \ref{tab:cnn width}. Expanding the results from \cite{morcos2018insights}, Figure \ref{fig:all_cnn} provides some evidence that better performing networks converged to similar solutions across the layers, however the 6 layer CNN showed this is not always the case. The poorer performance of the 6 layer CNN could be attributed to an over-fitting issue and to investigate this further, the potential memorisation within the neural representations would be need to be undertaken. These results can be expanded to further develop and explore better architecture solutions for End-to-End ASR performance, whilst gaining some insight of the effect architecture changes have upon network dynamics.

\section{Conclusion}

A comparative analysis of SVCCA and CKA has been undertaken for an End-to-End task and pathological components have been identified in CNN and Transformer models.
Further investigation of the attributes for the pathology would be required, for instance, do the unstable deeper layer neural representation correlations correspond to noisy components within ASR task? Furthermore, an extension to this work could be the analysis of neural representations on out of domain data, as the structural properties of the different layers could be beneficial to building models for few-shot-learning in ASR.
\section{Acknowledgements}
This work was partly supported by Voicebase Inc. at the Voicebase Research Center.

\bibliographystyle{IEEEtran}

\bibliography{paper1}

\end{document}